\documentclass[aps,pre,twocolumn,nofootinbib,tightenlines,superscriptaddress,nopacs,amsmath,amssymb,final,letterpaper]{revtex4-1}

\usepackage[utf8]{inputenc}
\usepackage{calc}
\usepackage{graphicx}
\usepackage{amsmath,amssymb,amsthm}
\usepackage{bbold}
\usepackage{bm}
\usepackage{color}
\usepackage[dvipsnames]{xcolor}
\usepackage{enumitem}
\usepackage{tikz}
\usepackage{float}
\usepackage[percent]{overpic}

\DeclareMathOperator*{\argmax}{arg\,max}
\DeclareMathOperator*{\argmin}{arg\,min}

\newcommand{\abs}[1]{\vert #1\vert}

\def\multiset#1#2{\ensuremath{\left(\kern-.3em\left(\genfrac{}{}{0pt}{}{#1}{#2}\right)\kern-.3em\right)}}

\usepackage[colorlinks=true,linkcolor=blue,urlcolor=blue,citecolor=blue,anchorcolor=blue]{hyperref}

\begin{document}
\title{Scalable inference of spatial regions and temporal signatures from time series}

\author{Jiayu Weng}
\affiliation{Institute of Data Science, University of Hong Kong, Hong Kong SAR, China}

\author{Alec Kirkley}
\email{alec.w.kirkley@gmail.com}
\affiliation{School of Computing and Data Science, University of Hong Kong, Hong Kong SAR, China}
\affiliation{Department of Urban Planning and Design, University of Hong Kong, Hong Kong SAR, China}
\affiliation{Urban Systems Institute, University of Hong Kong, Hong Kong SAR, China}

\begin{abstract}
Regionalization aims to partition a spatial domain into contiguous regions that share similar characteristics, enabling more effective spatial analysis, policy making, and resource management.
Existing approaches for spatial regionalization typically rely on static spatial snapshots rather than evolving time series. Meanwhile, most time series clustering methods ignore spatial structure or enforce spatial continuity through ad hoc regularization, constraining the number of inferred regions a priori either explicitly or implicitly.
Utilizing the minimum description length principle from information theory, here we propose an efficient and fully nonparametric framework for the regionalization of spatial time series. Our method jointly infers a spatial partition along with a set of representative time series archetypes (``drivers'') that best compress a spatiotemporal dataset, with a runtime log-linear in the number of time series.
We demonstrate that this method can accurately recover planted regional structure and drivers in synthetic time series, and can extract meaningful structural regularities in large-scale empirical air quality and vegetation index records. Our method provides a principled and scalable framework for spatially contiguous partitioning, allowing interpretable temporal patterns and homogeneous regions to emerge directly from the data itself.
\end{abstract}


\maketitle

\section{Introduction}

Regionalization~\cite{duque2007supervised,aydin2021quantitative} refers to the task of dividing a spatial domain into geographically contiguous regions that exhibit similar characteristics. Such partitions provide a simplified representation of complex spatial systems and are widely used in applications such as air-quality management~\cite{qiu2023airChina}, climate zoning~\cite{fovell1993climate, carvalho2016regionalization}, privacy preserving spatial aggregation~\cite{mu2015place,spielman2015reducing}, urban boundary delineation~\cite{morel2024bayesian}, and the study of socio-economic~\cite{assunccao2006efficient,li2014pcompact}, cultural~\cite{zhang2021mining}, and ecological~\cite{legendre1989spatial} spatial structure. By grouping spatial units into internally consistent regions, regionalization helps uncover large-scale spatial organization that may not be apparent from local observations~\cite{wolf2021spatially, poorthuis2018draw}.

Most existing regionalization methods are designed for static spatial attributes, where each location is represented by a single measurement or by variables aggregated over time~\cite{wei2021efficient, zhang2024advancing}. Classical approaches include the p-regions problem~\cite{duque2011p}, which seeks contiguous regions that maximize within-region homogeneity, along with spatially constrained clustering algorithms such as the automated zoning procedure (AZP)~\cite{openshaw1977geographical}, SKATER~\cite{assunccao2006efficient}, and REDCAP~\cite{Guo2008REDCAP}. While effective in many settings, these methods do not directly address the temporal dimension that is increasingly present in modern spatial datasets~\cite{shekhar2015spatiotemporal}. Examples include air-quality monitoring networks, remotely sensed vegetation indices tracking ecosystem dynamics, and long-term climate records such as temperature or precipitation. In such systems, functional similarity is better captured by shared temporal trajectories, seasonal cycles, or dynamical responses over time than by the resemblance of static structural features. 

At the same time, existing work on time series clustering has largely focused on non-spatial data~\cite{aghabozorgi2015time}. While non-spatial clustering methods---including methods for network community detection \cite{fortunato2010community}, spectral clustering \cite{ng2001spectral}, and time series clustering \cite{aghabozorgi2015time} among many other approaches \cite{gan2020data}---can sometimes provide good results for spatial boundary inference \cite{shen2019delineating,kirkley2020information}, these ``implicit'' regionalization methods often require extensive postprocessing to construct spatially contiguous regions in practice \cite{,morel2024bayesian}. In contrast, by directly enforcing spatial contiguity during the clustering process, ``explicit'' regionalization methods can be applied for this task with no postprocessing for contiguity \cite{aydin2021quantitative}. Some approaches for time series clustering~\cite{coppi2010fuzzy, yuan2015constrained} incorporate spatial information through additional regularization terms or spatial smoothing constraints, but these typically require tuning hyperparameters that control the strength of spatial coupling (and, ultimately, the structure of the clusters inferred).

Spatiotemporal clustering has been applied to a wide variety of data types, including point event data, moving objects, origin-destination trajectories, and spatially resolved time series measurements \cite{birant2007st-dbscan,shi2019spatiotemporal,ansari2020spatiotemporal}. Existing methods focusing on geo-referenced time series include density-based methods such as ST-DBSCAN \cite{birant2007st-dbscan}, the correlation-based method CorClustST \cite{husch2020corclustst}, hierarchical clustering approaches \cite{wang2023multivariate}, and process-oriented geographic boundary models \cite{zhang2024advancing}. While flexible for capturing different spatiotemporal regularities, these approaches typically rely on heuristic definitions of distance, density, correlation, or time-series similarity. As a result, the inferred regions and their interpretation can be sensitive to methodological choices and it is challenging to make consistent comparisons across datasets or regional contexts. Rigorous and conceptually coherent data summarization and cross-dataset comparisons require regionalization methods for spatial time series data to be derived from fundamental scientific principles. Another common limitation is the need to specify the number of regions a priori \cite{duque2007supervised, aydin2021quantitative,folch2014identifying}. Even when not fixed explicitly, the number of regions is usually determined indirectly through user-chosen parameters such as the density and distance thresholds in ST-DBSCAN \cite{birant2007st-dbscan} or the spatial correlation parameter in CorClustST \cite{husch2020corclustst}. In practice, the appropriate number of regions is rarely known in advance and may vary with temporal resolution, noise level, and time series length. Fixing the number of regions a priori may therefore result in spurious mesoscopic structure in noisy data or obscure heterogeneity at small spatial scales, while also enabling aesthetic and scientific confirmation biases. 

The minimum description length (MDL) principle~\cite{rissanen1978modeling, grunwald2007minimum} provides a natural basis for data clustering as it seeks models that achieve the shortest overall description of the data by balancing model complexity against goodness of fit.
The MDL principle has been applied in a variety of statistical modeling and unsupervised learning tasks, including the clustering of categorical data~\cite{li2004entropy}, real-valued vector data~\cite{georgieva2011cluster}, and time series in an aspatial setting~\cite{lu2010mdl,rakthanmanon2012mdl}. In the spatial domain, MDL has been employed to compress feature locations \citep{papadimitriou2005parameter}, perform density estimation \citep{yang2023unsupervised}, and identify spatial coverings of different cell classifications \cite{lakshmanan2002generalized}. The MDL principle has also been applied to extract a variety of structural regularities in network data, including communities \citep{peixoto2019bayesian}, core-periphery and hub structures \citep{gallagher2021clarified,kirkley2024identifying}, node hierarchies \cite{peixoto2022ordered,morel2025estimation}, and structural backbones \citep{kirkley2025fast}. The key strength of MDL is its ability to generate fully nonparametric summaries of data without requiring application-specific tuning, which allows for direct and straightforward comparisons across different contexts and datasets. 

While the MDL principle has been applied for regionalizing static demographic \cite{kirkley2022spatial} and mobility data \cite{morel2024bayesian}, its extension to spatiotemporal data poses a fundamentally harder challenge for interpretability and scalability by shifting from static attributes to potentially long and nonlinear temporal sequences. To accommodate this problem context, it is useful to infer representative structures within each cluster in addition to the clusters themselves. This strategy has been applied to cluster networks  \cite{coupette2022differentially,kirkley2023compressing} and their partitions \cite{kirkley2022representative,peixoto2021revealing}, both of which are unwieldy data structures, for easier cluster interpretation using the MDL principle. 

Here we propose a principled nonparametric framework for regionalizing spatial time series data based on the MDL principle. Our codelength objective allows us to partition spatial locations into contiguous regions, each with a representative driver time series, such that the total description length of the dataset is minimized when each time series is encoded conditionally on its regional driver. To guarantee contiguous regions, we model locations as nodes in a geographic adjacency network and perform a fast agglomerative merging process starting from singleton clusters, preserving network structure throughout the clustering. Our method is highly efficient, with a runtime that is log-linear in the number of 
network nodes (i.e. time series) and empirically  scaling easily to regionalize hundreds of thousands of time series. In synthetic experiments, we generate spatial time series with planted regional structure and demonstrate that our MDL-based method accurately recovers both the underlying regions and their driver patterns across a wide range of noise levels and parameter settings. For real-world evaluation, we apply the framework to two contrasting environmental case studies: daily air quality index (AQI) categories across California, and normalized difference vegetation index (NDVI) time series over Hong Kong. The AQI application reveals how the method uncovers elongated pollution regions that align with topography and known emission patterns, while the NDVI application identifies vegetation regions with distinct seasonal cycles that reflect urban–rural variation.


\section{Methods}
\label{sec:methods}

\begin{figure*}[!ht]
    \centering
    \includegraphics[width=\textwidth]{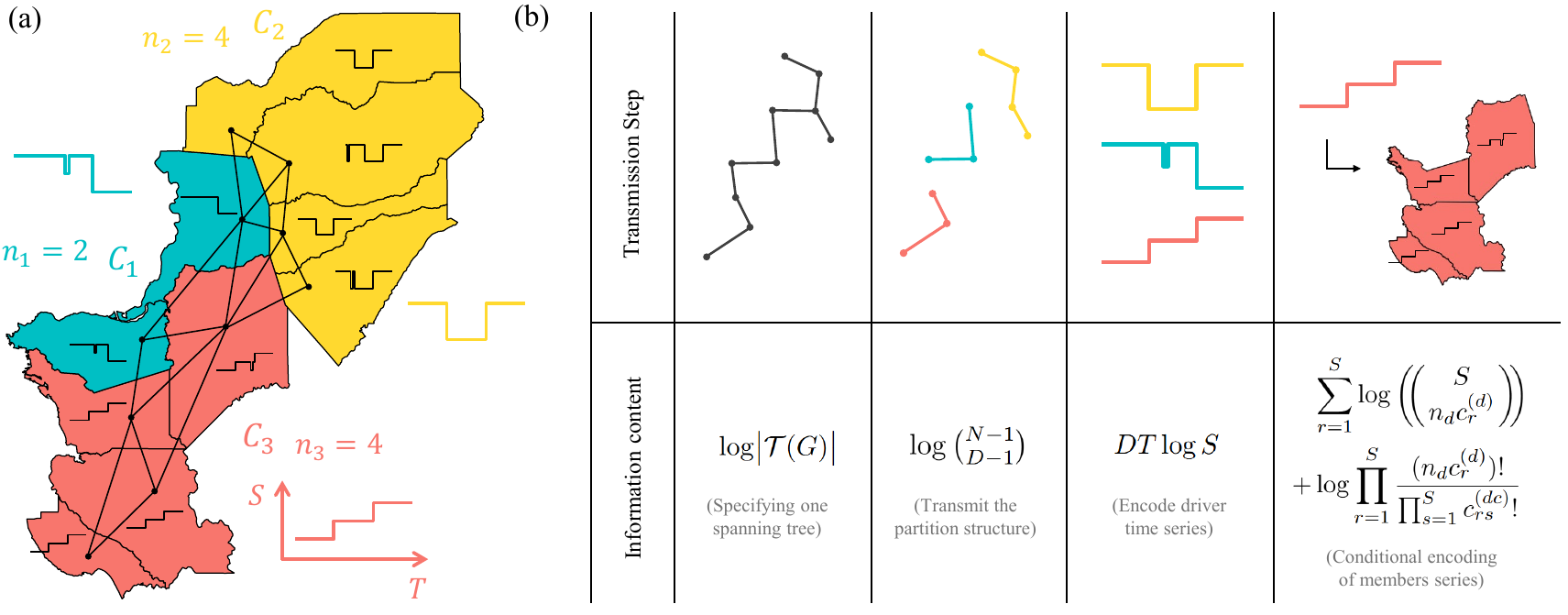}
    \caption{
    \textbf{Schematic illustration of the MDL formulation for spatiotemporal regionalization.}
    (a) Variables entering the description length objective for a candidate contiguous partition $\mathcal{C} = \{C_1, \dots, C_D\}$ of a spatial adjacency network of $N=10$ time series. The spatial units containing the time series are grouped into contiguous regions $C_d$ (colored), each associated with a representative driver time series $\bm{\zeta}(C_d)$ and $n_d$ unit-level time series $\{\bm{z}_i:i\in C_d\}$ that are encoded relative to their regional driver $\bm{\zeta}(C_d)$.
    (b) The total description length is decomposed into a sequence of transmission steps whose total sum is to be minimized under the MDL principle to identify the optimal partition $\mathcal{C}$ and drivers. From left to right, these steps include: specifying a spanning tree of the adjacency graph; selecting a contiguous partition by cutting the tree; encoding the driver time series for each region; and conditionally encoding the time series in each cluster $C_d$ based on the driver $\bm{\zeta}(C_d)$. The expressions shown in the lower row indicate the corresponding information content required for each step, according to Eq.~\ref{eq:MDL_total}.
    }
    \label{fig:method}
\end{figure*}

\subsection{Inference objective}
\label{sec:method-MDL}

We consider a set of $N$ spatial locations, each associated with a univariate time series over $T$ discrete time steps. The $N$ locations are treated as nodes in a spatial adjacency network $G = (V, E)$, where edges $(i,j) \in E$ connect neighboring locations based on spatial proximity. When natural spatial units exist (e.g. grid cells, census tracts, etc), we place the edge $(i,j)\in E$ when units $i$ and $j$ share some length of common border. Meanwhile, when the time series are located at isolated points in space, we place an edge $(i,j) \in E$ if the Voronoi cells of $i$ and $j$ share a boundary. Alternative constructions, such as $k$-nearest neighbor graphs, are also applicable within our framework as long as the resulting network reflects local proximity for assessing spatial contiguity of the partitions we will infer. 

Let $X \in \mathbb{R}^{N \times T}$ denote the raw time series data, with each entry $X_{it}$ representing a scalar value at location $i$ and time $t$. Before applying our method, $\bm{X}$ is discretized to form the categorical matrix $Z \in \{1,\dots,S\}^{N \times T}$, where $S$ denotes the size of the support of $\bm{Z}$ (e.g., $S=2$ for binary values), and prior to this discretization $\bm{X}$ may also be transformed using other time series preprocessing techniques (e.g. various detrending methods) if deemed relevant for the application. In this way all preprocessing choices are made explicitly at the beginning by the modeler based on domain expertise, with the discretized time series data $\bm{Z}$ permitting fully nonparametric MDL inference based on lossless compression thereafter. (It is not possible to perform any MDL analyses for continuous data without specifying at least one discretization parameter \cite{grunwald2007minimum}.) When the observations $\bm{X}$ are already categorical or ordinal, we can simply set $\bm{Z}=\bm{X}$, while for continuous data one can preprocess $\bm{X}$ so that the categories in $\bm{Z}$ reflect meaningful time series movements, such as increases or decreases in a binarization of the series. For a given discretization method, the inferred spatial partitions and compression identified under our method both converge as the number of bins $S$ increases, suggesting that in practice one should choose $S$ to be as large as possible while maintaining sufficient sampling density. We discuss further the effect of binning in Sec.~\ref{sec:results}.  

We can now compute the codelength (description length) $\mathcal{L}(\bm{Z},\mathcal{C})$ required to transmit the dataset $\bm{Z}$ to a receiver using a given partition $\mathcal{C}$ of the time series. For improved interpretability and coarse-graining for downstream analyses, each cluster $C_d$ is assigned a ``driver'' time series $\bm{\zeta}(C_d)$ representing the aggregate structure of the series within the cluster $C_d$. This driver time series $\bm{\zeta}(C_d)$ is used to aid in the compression of the time series $\bm{z}_i,~i\in C_d$ within its cluster. If the partition $\mathcal{C}$ and associated drivers capture substantial regularities in the time series data, then they will provide compression of $\bm{Z}$ (i.e., achieve a small codelength for transmission). If $\mathcal{C}$ has too few clusters and/or does not adequately summarize the heterogeneity in $\bm{Z}$, then the codelength will be high due to $\mathcal{C}$ poorly constraining the data $\bm{Z}$. On the other hand, if $\mathcal{C}$ has too many clusters, then the codelength will be high due to the complexity of transmitting $\mathcal{C}$. The optimal balance can be achieved under the MDL principle by minimizing $\mathcal{L}(\bm{Z},\mathcal{C})$ across possible contiguous partitions $\mathcal{C}$ of the time series $\bm{Z}$ and the possible latent drivers within each cluster. 

We will first assume that the receiver knows the number of time series $N$, the number of time periods $T$ in each time series $\bm{z}_i$ (where $\bm{z}_i=\{Z_{it}\}_{t=1}^{T}$), the support of $\bm{Z}$ with $S$ unique symbols, and the underlying spatial structure of the system (i.e. the node positions and adjacency graph $G$). We let the receiver also know the number of clusters (drivers) $D$ ahead of time for the construction of the codelength, noting that this quantity is still inferred automatically when we minimize the codelength over partitions $\mathcal{C}$. The information required to specify the global quantities $\{N,T,S,D\}$ contributes a very small constant to the description length that does not depend on the partition $\mathcal{C}$, so it can be safely ignored. Meanwhile, the spatial structure of the problem and $G$ are assumed known by the receiver since these constant factors also do not impact inference and their specification is considered out of scope for the MDL objective. 

To construct the description length, we will use a sequence of fixed-length codes equivalent to a hierarchical uniform Bayesian prior  \cite{mackay2003information}. Such a codelength is valid because according to the Kraft inequality, given any properly normalized probability distribution $P(x)$ over quantities $x\in \mathcal{X}$ in a finite set $\mathcal{X}$, there exists a valid (i.e., uniquely decodable) code in which $x$ is encoded with $-\log P(x)$ bits. (We let $\log\equiv \log_2$ for brevity.) Letting $P(x)=\abs{\mathcal{X}}^{-1}$ be the uniform distribution, we then have that $\log \abs{\mathcal{X}}$ is a valid codelength for $x$. Such fixed-length codes have been widely used in MDL applications for network \cite{peixoto2017nonparametric} and time series data \cite{kirkley2025transfer} among other discrete data types.   

We can encode the spatial structure of the partition $\mathcal{C}$ by first specifying a spanning tree of the underlying spatial adjacency graph $G$ among the $N$ time series, then specifying a cut of this spanning tree \cite{come2024bayesian}. Let $\mathcal{T}(G)$ denote the set of spanning trees of the adjacency graph. In a fixed-length code as described above, specifying one of these spanning trees requires $\log |\mathcal{T}(G)|$ bits. While this term is constant across partitions $\mathcal{C}$, it is included here as it contributes a non-negligible quantity to the total description length and in principle could be modified to produce a more efficient encoding if one could enumerate the number of spanning trees that allow for specific spatial cluster sizes. Given a spanning tree of the adjacency graph, any set of $D-1$ edge removals among the $N-1$ edges present in the tree defines $D$ spatially contiguous clusters, and since there are ${N-1\choose D-1}$ such cuts that could be made, it requires $\log \binom{N-1}{D-1}$ bits to specify the partition $\mathcal{C}$ once the spanning tree is known. Thus, the entire cost of transmitting the partition structure $\mathcal{C}$ is 
\begin{equation}
\label{eq:L_partition}
\mathcal{L}_{\text{part}}(\mathcal{C})=\log |\mathcal{T}(G)|+\log {N-1\choose D-1}.
\end{equation}

Next, we transmit the driver time series associated with each cluster $C_d\in \mathcal{C}$. Each driver consists of $T$ symbols drawn from an alphabet of size $S$, requiring a codelength of
\begin{equation}
\label{eq:L_driver}
\mathcal{L}_{\text{drv}}(\mathcal{C}) = D\,T\,\log S
\end{equation}
bits for all $D$ drivers. We will show in Sec.~\ref{sec:optimization} how to construct a locally optimal driver $\bm{\zeta}(C_d)$ directly from the cluster $C_d$, so we do not consider $\bm{\zeta}$ a separate free parameter in the inference process. The total description length for the cluster-level information is thus
\begin{align}
\mathcal{L}(\mathcal{C}) = \mathcal{L}_{\text{part}}(\mathcal{C}) + \mathcal{L}_{\text{drv}}(\mathcal{C}).   
\end{align}

Given $\mathcal{C}$ and the set of drivers, the remaining task is to transmit the member series within each cluster using a conditional encoding that exploits their similarity to the driver. The key object for compression is then the contingency table between the driver and its cluster of time series, which stores the joint distribution among the time series and allows for compression via nonlinear dependencies as in conditional entropy-based measures of similarity ~\cite{jerdee2024mutual, kirkley2022representative}. To construct this code, we concatenate $n_d$ copies of the driver sequence $\bm{\zeta}(C_d)$ into a single sequence $\bm{a}(C_d)=[\bm{\zeta}(C_d)\vert\vert \bm{\zeta}(C_d)\vert\vert \cdots \vert\vert \bm{\zeta}(C_d)]$ of length $n_d\times T$. We then use $\bm{a}(C_d)$ to transmit the $n_d$ observed time series in the cluster $C_d$, stored as a length $n_d\times T$ vector $\bm{b}(C_d)=[\bm{z}_{i_1}\vert\vert\bm{z}_{i_2}\vert\vert \cdots \vert\vert\bm{z}_{i_{n_d}}]$ ordered by increasing node index $i_k\in C_d$. Since each time series has length $T$ and the receiver knows $G$, this is sufficient to fully specify the time series within the cluster to the receiver. 

We define the contingency table $\bm{c}^{(dc)}$ with entries
\begin{align}
c^{(dc)}_{rs} = \sum_{t=1}^{T}\delta(\zeta_{t}(C_d),r)\sum_{i\in C_d}\delta(Z_{it},s),    
\end{align}
where $\zeta_{t}(C_d)$ is the $t$-th entry of the driver time series $\bm{\zeta}(C_d)$ and $\delta(\cdot,\cdot)$ is the Kronecker delta function. $c^{(dc)}_{rs}$ counts the number of times the time series within the cluster $C_d$ take the value $s$ at the same time the driver takes the value $r$. We now note that
\begin{align}
c_r^{(d)} &=\sum_{t=1}^{T}\delta(\zeta_{t}(C_d),r)\\
&=\frac{1}{n_d}\sum_{s=1}^{S}c^{(dc)}_{rs},   
\end{align}
which is the marginal count of the number of times symbol $r$ appears in $\bm{\zeta}(C_d)$, is fixed and known by the receiver after we transmit $\bm{\zeta}(C_d)$. These marginal counts can then be used to more efficiently transmit each row of the contingency table $\bm{c}^{(cd)}$, which must be specified to the receiver prior to transmitting the time series values $\bm{b}(C_d)$ within the cluster from the driver information $\bm{a}(C_d)$. There are $S$ columns in the $r$-th row $\{c^{(dc)}_{rs}\}_{s=1}^{S}$ of the contingency table $\bm{c}^{(dc)}$ which must be non-negative and sum to the correct marginal count $n_dc_r^{(d)}$. There are $\multiset{S}{n_dc_r^{(d)}}$ unique ways these entries can be configured, where
\begin{align}
\multiset{n}{k} = {n+k-1\choose k}    
\end{align}
is the multiset coefficient counting the number of unique ways to select $k$ elements from a dictionary of size $n$. We therefore require
\begin{align}
\mathcal{L}(\bm{c}^{(dc)}\vert C_d) = \sum_{r=1}^{S}\log \multiset{S}{n_dc_r^{(d)}}    
\end{align}
bits to transmit all rows $r$ of the contingency table $\bm{c}^{(dc)}$. 

Once the contingency table $\bm{c}^{(dc)}$ is known by the receiver, there are 
\begin{align}
{n_dc_r^{(d)} \choose c_{r,1}^{(dc)},~c_{r,2}^{(dc)},\cdots,c_{r,S}^{(dc)}} = \frac{(n_dc_r^{(d)})!}{\prod_{s=1}^{S}c_{rs}^{(dc)}!}    
\end{align}
ways to configure the values of the concatenated time series $\bm{b}(C_d)$ corresponding to the positions where the concatenated driver series $\bm{a}(C_d)$ takes the value $r$. Taking the logarithm and summing over all symbols $r=1,...,S$, it takes
\begin{align}
\label{eq:LZd}
\mathcal{L}(\bm{Z}^{(d)}\vert C_d,\bm{c}^{(dc)}) &= \sum_{r=1}^{S}\log {n_dc_r^{(d)} \choose c_{r,1}^{(dc)},~c_{r,2}^{(dc)},\cdots,c_{r,S}^{(dc)}} \\
&= \log \prod_{r=1}^{S}\frac{(n_dc_r^{(d)})!}{\prod_{s=1}^{S}c_{rs}^{(dc)}!}    
\end{align}
bits to transmit the time series $\bm{Z}^{(d)}=\{\bm{z}_i\}_{i\in C_d}$ within the cluster $C_d$ from the driver $\bm{\zeta}(C_d)$ once the joint distribution $\bm{c}^{(dc)}$ is known. Applying Stirling's approximation $\log x! \approx x\log x - x/\ln(2)$ to this combinatorial result, we can see that
\begin{align}
\mathcal{L}(\bm{Z}^{(d)}\vert C_d,\bm{c}^{(dc)}) \approx n_dT\times H(\bm{b}(C_d)\vert \bm{a}(C_d)),  \end{align}
where
\begin{align}
H(\bm{b}(C_d)\vert \bm{a}(C_d)) = -\sum_{r,s=1}^{S}p_{rs}^{(dc)}\log \frac{p^{(dc)}_{rs}}{p^{(d)}_{r}}    
\end{align}
is the standard Shannon conditional entropy of the concatenated time series $\bm{b}(C_d)$ within the cluster given the driver information $\bm{a}(C_d)$. Here, $p_{rs}^{(dc)}=\frac{c_{rs}^{(dc)}}{n_dT}$ and $p_{r}^{(d)}=\frac{c_{r}^{(d)}}{T}$ are the joint and marginal symbol probabilities extracted from the contingency table $\bm{c}^{(dc)}$. This suggests an approximate correspondence between the proposed objective and conditional entropy-based formulations of clustering \cite{kirkley2022representative}, but with built-in principled regularization provided by the transmission of the cluster-level information (including the drivers and contingency tables). 

Combining all terms and summing across clusters, the description length of the data $\bm{Z}$ given the partition $\mathcal{C}$ is
\begin{align}
\label{eq:Lcond}
\mathcal{L}(\bm{Z}\vert \mathcal{C}) = \sum_{d=1}^{D}[\mathcal{L}(\bm{c}^{(dc)}\vert C_d)+\mathcal{L}(\bm{Z}^{(d)}\vert C_d,\bm{c}^{(dc)})],    
\end{align}
so that the total description length of $\bm{Z}$ under partition $\mathcal{C}$ is
\begin{align}
\label{eq:MDL_total}
\mathcal{L}(\bm{Z},\mathcal{C})
&=
\mathcal{L}(\mathcal{C})+\mathcal{L}(\bm{Z}\vert \mathcal{C}) \nonumber\\
&=
\log |\mathcal{T}(G)|
+ \log {N-1 \choose D-1}
+ D T \log S \\
&+\sum_{d=1}^{D}\sum_{r=1}^{S}
\left[
\log \multiset{S}{n_d c_r^{(d)}}
+
\log
\frac{(n_d c_r^{(d)})!}
{\prod_{s=1}^{S} c_{rs}^{(dc)}!}
\right]\nonumber.
\end{align}
Through the first three terms from $\mathcal{L}(\mathcal{C})$, Eq.~\ref{eq:MDL_total} provides a penalty for partitions $\mathcal{C}$ with a large number of clusters $D$, as more information is required to specify their finer spatial structure. Meanwhile, the sum from $\mathcal{L}(\bm{Z}\vert\mathcal{C})$ provides a penalty for partitions $\mathcal{C}$ that are not fine enough to capture structural regularities in $\bm{Z}$ across space. The optimal partition $\mathcal{C}_{\text{MDL}}$ under the MDL principle balances these competing effects by minimizing the description length of Eq.~\ref{eq:MDL_total}, thus
\begin{align}
\mathcal{C}_{\text{MDL}} = \argmin_{\mathcal{C}}\{\mathcal{L}(\bm{Z},\mathcal{C})\}.    
\end{align}

A schematic of this objective is shown in Fig.~\ref{fig:method} using example synthetic time series.

\subsection{Optimization and model selection}
\label{sec:optimization}

In order to minimize the description length $\mathcal{L}(\bm{Z},\mathcal{C})$ over spatial partitions $\mathcal{C}$, we must first determine a procedure to construct the MDL-optimal driver time series $\bm{\zeta}(C_d)$ from the time series $\bm{Z}^{(d)}$ that fall within the cluster $C_d$. Looking at Eq.~\ref{eq:Lcond}, we can observe that the optimal driver $\bm{\zeta}(C_d)$ for cluster $C_d$ will result in a contingency table $\bm{c}^{(cd)}$ that minimizes the cluster-level description length $\mathcal{L}(\bm{c}^{(dc)}\vert C_d)+\mathcal{L}(\bm{Z}^{(d)}\vert C_d,\bm{c}^{(dc)})$ given the data $\bm{Z}^{(d)}$. 

Defining the majority vote estimator
\begin{align}
\zeta_t(C_d) = \argmax_{s\in \{1,..,S\}}\left\{m_{ts}\right\},
\end{align}
where
\begin{align}
m_{ts} = \sum_{i\in C_d}\delta(z_{it},s)    
\end{align}
is the number of times the symbol $s$ occurs among the time series $\bm{Z}^{(d)}$ at time $t$, we can show that this estimator provides the locally optimal driver for sufficiently well-sampled time series $\bm{Z}^{(d)}$ and homogeneous clusters $C_d$ (see Appendix~\ref{app:driver_optimality}). This majority vote estimator simply assigns to the driver $\bm{\zeta}(C_d)$ the most frequently occurring symbol $s$ from the cluster time series $\bm{Z}^{(d)}$ at time $t$, allowing us to quickly update this driver when performing our agglomerative optimization scheme for $\mathcal{L}(\bm{Z},\mathcal{C})$.

Directly minimizing the MDL objective $\mathcal{L}(\bm{Z},\mathcal{C})$ over all contiguous partitions $\mathcal{C}$ is computationally intractable, so here we adopt an approximate greedy agglomerative merging strategy. (We compare greedy optimization with the exact solution on small synthetic datasets in Appendix Fig.~\ref{fig:compare_exact}, finding that the greedy algorithm provides perfectly optimal or near-optimal compression for all cases.)
The algorithm starts from the trivial partition in which each spatial location $i\in \{1,...,N\}$ forms its own region. At each step, we then consider all admissible merges between adjacent regions and select the merge that yields the largest decrease (or the smallest increase) in the total description length $\mathcal{L}(\bm{Z},\mathcal{C})$.

Two regions $C_{d}$ and $C_{d'}$ are considered adjacent if and only if there exists a pair of nodes $u \in C_d,v\in C_{d'}$ such that $(u,v)\in E$. We can quickly evaluate the change in description length produced by merging two adjacent clusters $C_d,C_{d'}$ as
\begin{align}
\Delta\mathcal{L}(d,d') &\propto \mathcal{L}(\bm{Z}^{(d)}\cup\bm{Z}^{d'}\vert C_{d}\cup C_{d'})\\
&-[\mathcal{L}(\bm{Z}^{(d)}\vert C_d)+\mathcal{L}(\bm{Z}^{(d')}\vert C_{d'})] \nonumber,
\end{align}
where we've removed terms that only depend on global constants and $D$, which can be updated only once at each step of the algorithm, before all new admissible merges are computed. During the merge proposals, we store the values $\{m_{ts}\},\{c^{(dc)}_{rs}\}$ and incrementally update these values, so that the driver $\bm{\zeta}(C_{d''})$ and the terms $\mathcal{L}(\bm{c}^{(d''c)}\vert C_{d''})+\mathcal{L}(\bm{Z}^{(d'')}\vert C_{d''},\bm{c}^{(d''c)})$ can be quickly updated for the new cluster $C_{d''}=C_d\cup C_{d'}$. 

Recording the description length at each merge step and merging until $\mathcal{C}$ contains all $N$ time series yields a full merge trajectory $\mathcal{L}^{(D)}_{\text{MDL}}$ giving the greedily-optimal partition into $D=N,N-1,...,1$ spatially contiguous clusters. The final regionalization is then chosen among these partitions as the partition attaining the minimum description length along this trajectory, or $\argmin_{D}\{\mathcal{L}^{(D)}_{\text{MDL}}\}$. In this way, the MDL objective performs model selection for the number of clusters $D$ automatically, and in our example applications, we therefore choose to let the description length tell us exactly how many clusters are in the data. However, in some applications it may be preferable to have a fixed value of $D$, and this can easily be accommodated in our algorithm by simply performing the greedy merge moves until the desired number of clusters $D$ is reached.

Starting from $N$ singleton regions, the algorithm performs $N-1$ merges. Since the adjacency graph is induced by 2D planar regions, the number of admissible region adjacencies scales linearly with the number of regions $N$, and these candidates $(d,d')$ are maintained in a priority queue ordered by $\Delta\mathcal{L}(d,d')$. Since insertion and removal in a binary heap require $O(\log N)$ time complexity when the queue contains $O(N)$ candidates, maintaining and updating the merge queue over all steps gives $O(N\log N)$ bookkeeping complexity. The total runtime also depends on updating cluster-level quantities, including driver sequences and contingency counts. The driver $\zeta(C_{d}\cup C_{d'})$ is computed by merging the counts $\{m_{ts}\}$ from each cluster, which can be stored in hash tables, and so we incur $O(T)$ time complexity to update all values $\zeta_t$ for $t=1,...,T$ using the majority vote estimator. The contingency tables can then be computed from the updated counts $\{m_{ts}\}$ with the same runtime scaling. The overall complexity of this optimization algorithm is then $O(TN\log N)$, which is only modestly higher than constructing the time series dataset $\bm{Z}$ itself, making the method highly scalable in practice. In Appendix~\ref{app:time_complexity}, we assess the empirical runtime scaling of the full algorithm to verify this theoretical complexity.

To summarize the quality of a regionalization, we can use the inverse compression ratio
\begin{equation}
\eta = \mathcal{L}_{\mathrm{MDL}}/\mathcal{L}_0,
\label{eq:compression}
\end{equation}
where $\mathcal{L}_{\mathrm{MDL}}$ is the description length of the selected MDL solution, and $\mathcal{L}_0$ is the description length of the unclustered baseline in which each spatial location forms its own region. Smaller values of $\eta$ indicate stronger compression and hence a more efficient representation of the data as spatially contiguous clusters summarized by driver time series.

\begin{figure*}[!ht]
    \centering
    \includegraphics[width=\textwidth]{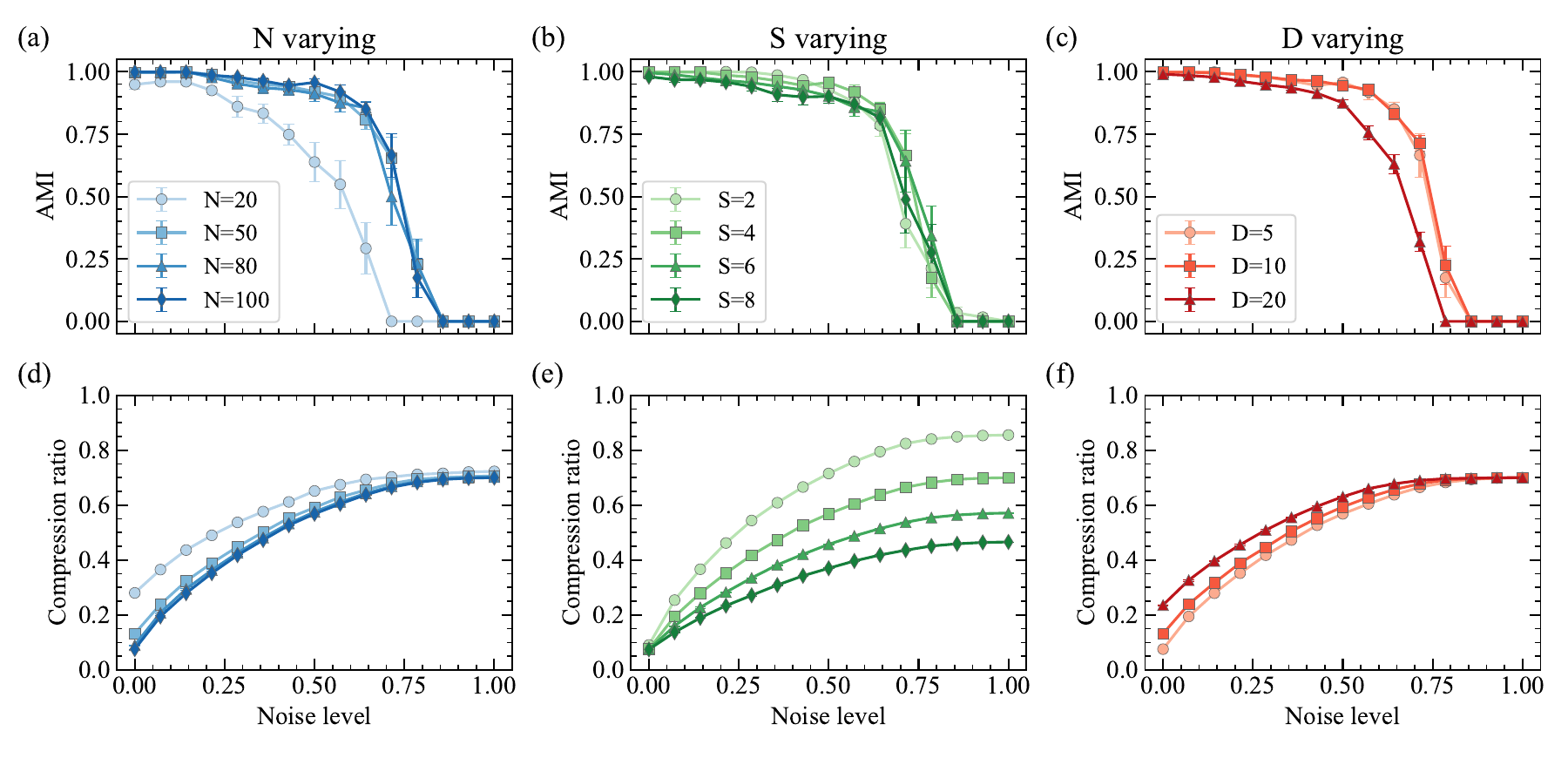}
    \caption{
    \textbf{Performance on synthetic datasets with planted cluster structure.} (a–c) Clustering accuracy, measured by adjusted mutual information (AMI) \cite{vinh2009information}, while varying (a) the number of nodes $N$, (b) the number of discrete states $S$, and (c) the number of clusters $D$.
    (d–f) Corresponding MDL compression ratios for the same settings. Each point shows the mean over 10 random realizations, with error bars indicating two standard errors.
    }
    \label{fig:synthetic}
\end{figure*}

\section{Results}
\label{sec:results}
We evaluate the proposed framework on both synthetic and real spatiotemporal datasets. We first examine its ability to recover planted spatial clusters and representative temporal patterns under varying noise levels in synthetic data. We then apply the method to two environmental case studies: daily air quality index (AQI) categories across California and NDVI-based vegetation dynamics over Hong Kong. Across these examples, the inferred partitions illustrate how the MDL framework balances spatial contiguity with temporal similarity to reveal meaningful spatiotemporal structure.

\subsection{Synthetic Data}
\label{sec:synthetic}
We first evaluate the performance of our regionalization framework on synthetic datasets with known planted structure. These controlled experiments allow us to systematically assess the ability of the method to recover spatially contiguous clusters and their associated driver time series under varying levels of problem complexity. 

The synthetic data generation procedure proceeds as follows. We begin by placing $N$ points uniformly at random in the plane and construct a spatial adjacency graph based on the Voronoi tessellation of these points (i.e. the Delaunay triangulation of the points \cite{de2010triangulations}). We then compute a minimum spanning tree (MST) on this graph and cut $D-1$ randomly chosen edges of the MST to obtain $D$ contiguous spatial clusters. 
Within each cluster, we select one location uniformly at random to serve as the driver. The driver time series is generated by independently drawing symbols from a discrete alphabet of size $S$ at each time step $t = 1, \dots, T$. For every non-driver node $i$, we start from an exact copy of its corresponding driver series $\bm{\zeta}(C_d)$. Noise is then introduced by independently replacing the symbol $z_{it}$ at each time step $t$ with a probability proportional to the noise level. When a random replacement occurs at time $t$, the original symbol is replaced by a value drawn uniformly at random. At each time step, a perturbation replaces the original value with one of the $S{-}1$ other states with probability $1-S^{-1}$, so the same symbol replacement probability leads to different effective corruption levels for different $S$. We therefore normalize the flip probability by $1-S^{-1}$ to obtain the noise parameter so that degradation in the algorithm performance occurs at similar levels of noise across values of $S$.

We systematically vary one of the parameters $N$, $S$, or $D$ while keeping the others fixed at $N=100$, $T=51$, $S=4$, and $D=5$. For each setting, we sweep the rescaled noise level from $0$ (perfect copies of the driver) to $1$ (maximally corrupted series) and generate ten independent realizations per noise level to reduce spurious fluctuations. The method of Sec.~\ref{sec:methods} is then applied to the synthetic dataset and clustering accuracy is evaluated using the adjusted mutual information (AMI) between the inferred partition and the planted cluster labels \cite{vinh2009information}. We also record the inverse compression ratio $\eta$ (Eq.~\ref{eq:compression}).

The results are summarized in Fig.~\ref{fig:synthetic}. Each data point represents the average over 10 independent realizations of the synthetic recovery experiment, with error bars indicating two standard errors. As expected, when the noise is small, the time series of non-driver locations remain close to their corresponding drivers, and the planted spatial partition can be recovered with high accuracy. As noise increases, AMI drops sharply once the temporal signal becomes dominated by random perturbations, with the transition occurring around a noise level of approximately $0.6$ in these experiments. 
We can see that, when fixing other parameters, smaller values of $N$ lead to earlier degradation in AMI (Fig.~\ref{fig:synthetic}(a)), while larger $N$ yields more stable performance over a broader range of noise levels, consistent with having more observations to support spatially constrained merging decisions. In Fig.~\ref{fig:synthetic}(b), the clustering accuracy shows relatively weak dependence on the number of discrete states $S$ after the noise rescaling, indicating that the method is not overly sensitive to the size of the discrete state space. Additionally, increasing the number of planted regions $D$ makes recovery more challenging (Fig.~\ref{fig:synthetic}(c)), as smaller regions contain fewer time series and provide less redundancy for the MDL objective to exploit.

Fig.~\ref{fig:synthetic}(d–f) show the corresponding MDL compression ratios $\eta$, which are used to select the optimal partition. For all parameter settings, $\eta$ increases smoothly with noise, reflecting the progressive loss of compressible temporal structure. Its close correspondence with AMI indicates that partitions recovered accurately by the algorithm are also those that yield the greatest compression gains. Larger values of $S$ lead to systematically lower compression ratios at low and moderate noise, because when the symbol space is larger, shared temporal structure provides greater savings in description length. Overall, the synthetic experiments show that the MDL objective both recovers the planted spatial structure and captures its progressive loss under increasing noise.

\subsection{Case Study: AQI in California}

\label{sec:AQI}  
\begin{figure*}[!ht]
    \centering
    \includegraphics[width=\textwidth]{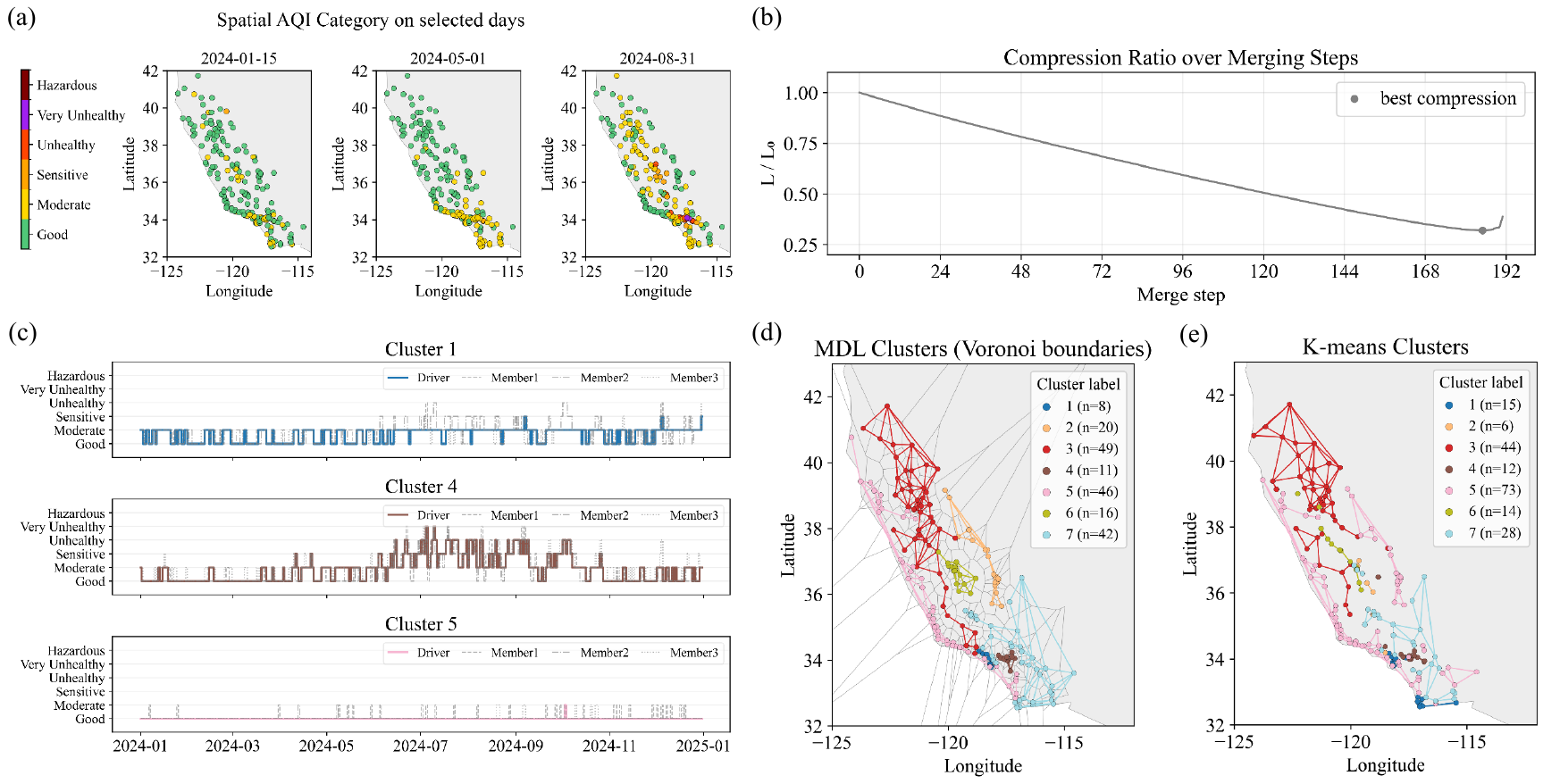}
    \caption{
    \textbf{Daily air quality index (AQI) time series across California.}
    (a) Spatial distribution of AQI categories on three selected dates in 2024.
    (b) MDL compression ratio $\eta$ at each agglomerative merging step, with the minimum highlighted as the best compression value used to obtain the clustering result.
    (c) Three member time series (grey) and their corresponding driver patterns (colored) inferred by the MDL-based method for three representative clusters.
    (d) Spatial clustering obtained by the proposed MDL-based method on a Voronoi-based spatial adjacency network. Light grey lines indicate Voronoi boundaries, points are colored by cluster assignment, and thin edges connecting nearby points show Voronoi adjacency links in the same cluster.
    (e) $K$-means clustering with the same number of clusters as in (d), shown for comparison with the same edge representation, where some clusters appear as disconnected components.
    }
    \label{fig:AQI}
\end{figure*}

For the first example with empirical spatiotemporal data, we study discrete-valued time series, the air quality index (AQI) across California in 2024. AQI is a standardized metric used to quantify air pollution levels and associated health risks, defined as the maximum value among multiple pollutant-specific indices~\cite{horn2024air}. AQI observations fall into one of six ordinal categories: good (0--50), moderate (51--100), unhealthy for sensitive groups (101--150), unhealthy (151--200), very unhealthy (201--300), and hazardous (301+) \cite{kanchan2015review}.

We obtain daily monitoring-station data from the U.S. Environmental Protection Agency (EPA) Outdoor Air Quality Data platform through its daily data download tool. The dataset covers six major pollutants in California for the year 2024: CO, NO$_2$, O$_3$, PM$_{2.5}$, PM$_{10}$, and SO$_2$. For each monitoring site and date, we define the overall AQI as the maximum among the six pollutant-specific AQI values, discretizing these into the six official categories above. To ensure temporal completeness, we retain only sites in which at least 80\% of the possible observations exist, and fill missing values using linear interpolation~\cite{lepot2017interpolation}. After preprocessing, the dataset consists of 192 monitoring sites with 366 daily observations per site. These series, along with the spatial coordinates of the monitoring stations, are used as inputs for subsequent clustering analysis.

Fig.~\ref{fig:AQI} summarizes the regionalization results. Fig.~\ref{fig:AQI}(a) shows the spatial distribution of AQI categories on three selected dates, revealing clear seasonal variation. On the winter date January 15, most monitoring stations remain in the lower AQI categories, while during the summer months, especially August, many areas exhibit higher AQI values. This seasonal difference is consistent with stronger secondary pollutant formation during warmer periods, particularly for ozone, whose production is often enhanced by higher temperatures and photochemical activity~\cite{bedsworth2012air, otero2021temperature}.
We report the MDL inverse compression ratio $\eta$ over successive agglomerative merging steps in Fig.~\ref{fig:AQI}(b), with the minimum indicating the selected clustering solution. The optimal partition attains $\eta \approx 0.32$, meaning that the data can be encoded in roughly one-third of the description length required by the un-clustered baseline.
In Fig.~\ref{fig:AQI}(c), we show representative time series for three inferred clusters, including the regional driver patterns and several member series within each cluster. These examples illustrate that the 
inferred regions are characterized by coherent temporal dynamics. 
Full driver time series and summary statistics for all clusters are provided in Appendix~\ref{app:details_main_text} and Fig.~\ref{fig:AQI_appendix}.
Fig.~\ref{fig:AQI}(d) shows that the MDL solution forms spatially contiguous regions on the Voronoi-based adjacency network, without isolated assignments. The resulting clusters closely follow California’s major topographic features, such as coastal corridors, inland valleys, and basin structures.
The regionalization also aligns well with known air-quality regions. Cluster~4 corresponds closely to the South Coast Air Basin, while Cluster~6 mainly covers the San Joaquin Valley. These two regions have historically been among the most polluted regions in the United States~\cite{bedsworth2012air}. Consistent with this broader understanding, the driver series of Cluster~4 shows the highest summertime AQI levels among all seven clusters, while Cluster~6 forms a distinct inland region separated from the coastal clusters.
In contrast, coastal regions such as Cluster~5, as well as parts of Cluster~2, maintain generally better air quality throughout the year. Cluster~5 lies near the Pacific coast, reflecting the moderating influence of maritime airflow. These patterns match the broad structure of California’s air basins and the spatial organization of pollution sources and meteorology.

Fig.~\ref{fig:AQI}(e) shows the spatial organization by $k$-means using the same number of clusters found by MDL. The number of clusters is fixed to this value since $k$-means (as well as other comparable time series clustering methods \cite{gan2020data}) require free parameters that either directly or indirectly fix the number of clusters. By comparison, $k$-means produces visibly more fragmented and interleaved clusters, mixing stations that are geographically separated and yielding less coherent regional shapes. Taken together, Fig.~\ref{fig:AQI} shows that the MDL criterion selects a partition that simultaneously (i) achieves strong compression, (ii) yields interpretable regional driver patterns, and (iii) respects spatial contiguity in a way that generic feature-space clustering does not. 
We provide a longitudinal comparison of the California AQI regionalization results for 2000, 2010, and 2020 in the Appendix~\ref{app:longitudinal_AQI} and Fig.~\ref{fig:AQI_different_years}.

\subsection{Case Study: NDVI in Hong Kong}

\label{sec:NDVI}
\begin{figure*}[!ht]
    \centering
    \includegraphics[width=\textwidth]{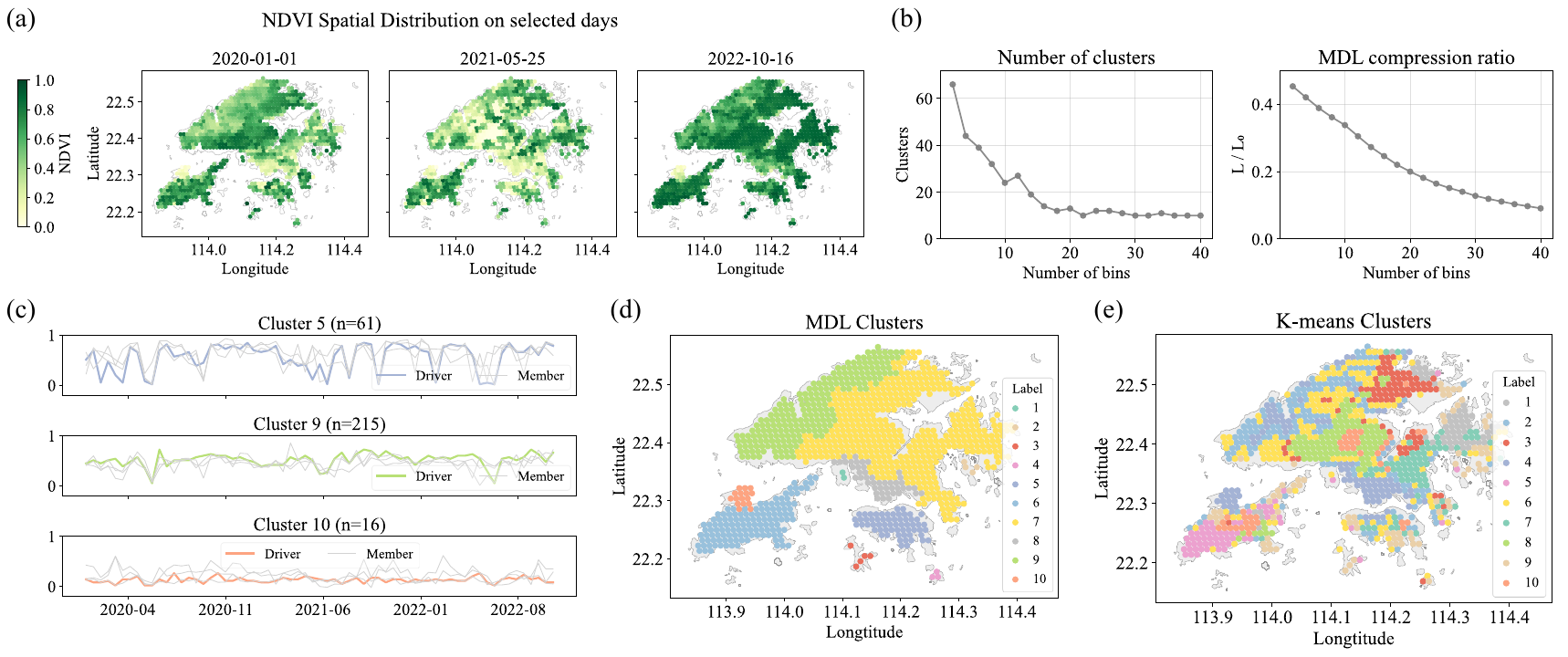}
    \caption{
    \textbf{16-day NDVI time series across Hong Kong.}
    (a) Spatial distribution of normalized difference vegetation index (NDVI) on three selected dates from 2020 to 2022; colors indicate NDVI magnitude.
    (b) Effect of the time series discretization resolution on the resulting number of clusters (left) and on the MDL compression ratio $\eta$ (right), showing that both stabilize as the resolution increases.
    (c) Time series of three members (grey) and driver patterns (colored) obtained from MDL-based method for three representative clusters.
    (d) Spatial clustering obtained by the MDL method (with $S=40$), using Voronoi adjacency to maintain spatial contiguity. Points are again colored by cluster assignment.
    (e) $K$-means clustering with the same number of clusters as in (d), shown for comparison.
    }
    \label{fig:NDVI}
\end{figure*}

We also test our method on continuous-valued time series, analyzing vegetation dynamics across Hong Kong during 2020--2022 using the normalized difference vegetation index (NDVI). NDVI is a widely used vegetation indicator derived from red and near-infrared reflectance, ranging from $-1$ to $1$, with higher values indicating greater vegetation density and health~\cite{pettorelli2005using, fung2001study}. We use the MODIS MOD13A2 product from NASA, accessed through Google Earth Engine, which provides 16-day composite NDVI data at 1 km spatial resolution.
To ensure data quality, we again exclude all spatial locations with more than 20\% of observations missing across the study period, linearly interpolating the remaining time series to fill in the gaps. After preprocessing, we obtain $N=939$ spatial locations with $T=65$ observations per location. 
We then discretize the continuous-valued NDVI matrix $\bm{X} \in \mathbb{R}^{N \times T}$ into a discrete-valued matrix $\bm{Z} \in \{0, 1, \dots, S-1\}^{N \times T}$ using 
\begin{align}
z_{it} = \mathrm{round}\left( \frac{X_{it} - x_{\min}}{x_{\max} - x_{\min} + \varepsilon} \cdot (S - 1) \right),
\end{align}
where $x_{\min}$ and $x_{\max}$ denote the global minimum and maximum over all entries in $\bm{X}$, and $\varepsilon$ is a small constant added for numerical stability. This produces a symbolic spatiotemporal dataset suitable for our combinatorial description length calculations. We rerun the algorithm for all values of $S$ in the range $\{2,...,40\}$ to examine the effect of this discretization resolution.

Fig.~\ref{fig:NDVI} summarizes the regionalization results. The NDVI snapshots in Fig.~\ref{fig:NDVI}(a) provide a qualitative reference for spatial heterogeneity: persistently high NDVI appears in major mountainous and country-park areas, while the lowest NDVI concentrates in the dense urban cores. Fig.~\ref{fig:NDVI}(b) reports the inferred number of clusters (left) and the MDL compression ratio $\eta$ (right) as a function of the number of bins. For coarse discretizations, the method returns many small clusters and only moderate compression. As the resolution increases, both the cluster count and the compression ratio rapidly stabilize, so that beyond roughly 30 bins the number of regions plateaus while $\eta$ approaches its minimum value. This suggests that the MDL objective is not highly sensitive to the precise discretization level $S$ and that larger values of $S$ provide more stable results, so in practice $S$ should be chosen as large as possible without introducing too much sampling noise from sparse bins. The results for $S=40$, which yields $D=10$ clusters, are shown in Fig.~\ref{fig:NDVI}(c--e). Additional visualizations and summary statistics for all clusters are provided in Appendix~\ref{app:details_main_text} and Fig.~\ref{fig:NDVI_appendix}.

Fig.~\ref{fig:NDVI}(c) shows representative driver patterns together with several member series for three clusters, while the inferred spatial partition is shown in Fig.~\ref{fig:NDVI}(d). The MDL solution yields contiguous regions that align well with Hong Kong’s major landforms and land-use structure, particularly across the main islands and urbanized corridors. Clusters 1--4 consist primarily of fragmented small islands and sparsely sampled offshore locations, reflecting geographic isolation under the contiguity constraint. Cluster~5 corresponds to Hong Kong Island, forming a coherent island-scale region. Cluster~6 aligns with Lantau Island, capturing the distinct vegetation regime of the territory's largest island, while Cluster~10 captures the airport area as a specialized land-use zone with visibly lower NDVI values. On the mainland, Cluster~8 covers the high-density urban belt (including Kowloon and the Tsuen Wan corridor), characterized by systematically lower NDVI levels. In contrast, Cluster~7 spans the largest continuous vegetated areas in the New Territories, including major country-park and mountainous regions (e.g., the Sai Kung area and the Tai Mo Shan region), exhibiting consistently higher NDVI. Finally, Cluster~9 captures northern New Territories peri-urban and boundary-adjacent areas (e.g., Tin Shui Wai and Sheung Shui), which act as a transitional zone between the urban core and the most vegetated interior regions. Taken together, these clusters reveal a natural urban gradient and island-specific vegetation regimes emerging directly from the NDVI dynamics under spatial contiguity.

Fig.~\ref{fig:NDVI}(e) reports $k$-means clustering using the same number of clusters as in Fig.~\ref{fig:NDVI}(d). $K$-means produces visibly mixed labels across space, breaking up otherwise coherent regions. In contrast, the MDL partition jointly optimizes temporal coherence and spatial contiguity, yielding compact regions that align with known urban cores, major country parks, and transitional zones between them. 
To illustrate the scalability of the method beyond the Hong Kong case study, we also present additional large-scale NDVI applications in Guangdong Province and across mainland China in Appendix~\ref{app:large_NDVI} and Figs.~\ref{fig:NDVI_Guangdong}--\ref{fig:NDVI_China}. Together with the AQI example, these results demonstrate that the framework can extract interpretable spatial structure from both discrete and continuous environmental time series.


\section{Conclusion}

In this work, we introduce an efficient nonparametric framework for regionalizing spatial time series based on the minimum description length (MDL) principle. The method identifies spatially contiguous regions with shared temporal structure, while inferring representative driver patterns that summarize the dominant dynamics within each region. By formulating the problem as an information compression task, the framework balances model complexity against temporal homogeneity, allowing the number of regions to emerge directly from the data without requiring user-specified hyperparameters.
Experiments on synthetic benchmarks and empirical environmental data demonstrate that the approach recovers meaningful time series regularities while ensuring spatial contiguity, with the empirical results aligning with known environmental structures including air-quality basins in California and vegetation dynamics across Hong Kong.

Several directions for future work remain. First, the framework could be extended to incorporate multivariate time series, allowing multiple environmental or socio-economic variables to jointly inform regionalization. Second, alternative encoding schemes may better capture temporal dependencies beyond symbol-level similarity, such as lagged relationships or seasonal dynamics. Finally, applying the method to larger spatiotemporal datasets and additional application domains may further reveal its potential as a general tool for uncovering structure in spatial time series data. More broadly, an interesting direction is the detection of spatiotemporal events or clusters from data with both spatial locations and time stamps~\cite{belhadi2020space}. Such problems arise, for example, in the study of infectious disease outbreaks or extreme environmental events. Extending MDL-based approaches to these settings may provide a principled way to identify coherent space–time structures.


\section*{Acknowledgments}
\vspace{-\baselineskip}
This work was supported in part by the Hong Kong Research Grants Council through General Research Fund Project No. 17301024 and through the HKU-100 Start Up Fund (AK). The authors thank Yanting Zhang for helpful discussions.


\clearpage
\appendix
\onecolumngrid

\section{Local optimality of majority vote drivers}
\label{app:driver_optimality}

We want to prove that the majority vote estimator
\begin{align}   
\zeta_t(C_d) = \argmax_{s\in \{1,..,S\}}\left\{m_{ts}\right\}
\end{align}
for the driver $\bm{\zeta}(C_d)$ results in a contingency table $\bm{c}^{(dc)}$, defined as
\begin{align}
c^{(dc)}_{rs} = \sum_{t=1}^{T}\delta(\zeta_t,r)m_{ts},
\end{align}
that satisfies the solution to the local MDL problem
\begin{align}
  \argmin_{\bm{c}^{(dc)}\vert \bm{Z}^{(d)}}\left\{\sum_{r=1}^{S}
\left[
\log \multiset{S}{n_d c_r^{(d)}}
+
\log
\frac{(n_d c_r^{(d)})!}
{\prod_{s=1}^{S} c_{rs}^{(dc)}!}
\right]\right\}.
\end{align}
$m_{ts}$ is a fixed constraint in the optimization problem since we condition on $\bm{Z}^{(d)}$. Expanding the multiset coefficient and combining terms, we can simplify the objective to
\begin{align}
\argmin_{\bm{c}^{(dc)}\vert \bm{Z}^{(d)}}\left\{\sum_{r=1}^{S}
F(\bm{\kappa}_r)\right\},
\end{align}
where $\bm{\kappa}_r=\{c^{(dc)}_{rs}\}_{s=1}^{S}$ is the $r$-th row of $\bm{c}^{(dc)}$ and
\begin{align}
F(\bm{\kappa}_r)=\log\frac{(n_d c_r^{(d)}+S-1)!}
{(S-1)!\prod_{s=1}^{S} c_{rs}^{(dc)}!}.
\end{align}
We can thus optimize over each row $\bm{\kappa}_r$ of the contingency table separately.

We can now prove that the majority vote estimator is locally optimal for sufficiently concentrated counts (which are themselves a result of the majority vote estimator). Consider any local perturbation which perturbs the majority vote estimator $\zeta_t$ at time $t$ from its current value $\zeta_t=r_0$ to a new value $\zeta_t=r_1$. By construction of the estimator, we know that $m_{tr_0}>m_{tr_1}$. The change in the local description length is given by
\begin{align}
\Delta_t = [F(\bm{\kappa}_{r_0}-\bm{m}_t)+F(\bm{\kappa}_{r_1}+\bm{m}_t)]-[F(\bm{\kappa}_{r_0})+F(\bm{\kappa}_{r_1})],
\end{align}
where $\bm{m}_t=\{m_{ts}\}_{s=1}^{S}$.

Ignoring the constant $(S-1)!$ for now since it will cancel out anyway, we can expand $F$ for $c_r^{(d)}\gg 1$ as
\begin{align}
F(\bm{\kappa}_r) \approx \log\frac{(n_d c_r^{(d)})!}
{\prod_{s=1}^{S} c_{rs}^{(dc)}!} \approx n_d c_r^{(d)} H\left(\frac{\bm{\kappa}_r}{n_dc_r^{(d)}}\right),    
\end{align}
where $H$ is the Shannon entropy and we have applied a Stirling approximation in the second step. Substituting this in for $\Delta_t$, we have
\begin{align}
\Delta_t &\approx [n_dc^{(d)}_{r_0}-n_d]H\left(\frac{\bm{\kappa}_{r_0}-\bm{m}_t}{n_dc^{(d)}_{r_0}-n_d}\right)-n_dc^{(d)}_{r_0}H\left(\frac{\bm{\kappa}_{r_0}}{n_dc^{(d)}_{r_0}}\right)\\
&+[n_dc^{(d)}_{r_1}+n_d]H\left(\frac{\bm{\kappa}_{r_1}+\bm{m}_t}{n_dc^{(d)}_{r_1}+n_d}\right)
-n_dc^{(d)}_{r_1}H\left(\frac{\bm{\kappa}_{r_1}}{n_dc^{(d)}_{r_1}}\right).
\end{align}
For $\abs{\abs{\bm{\kappa}_{r}}}_1 \gg \abs{\abs{\bm{m}_t}}_1 $, or in other words, $c^{(d)}_{r} \gg 1$ as before, we have that this can be expanded as
\begin{align}
\Delta_t \approx n_d \times \left[H\left(\frac{\bm{\kappa}_{r_0}}{n_dc^{(d)}_{r_0}}\right)-H\left(\frac{\bm{\kappa}_{r_1}}{n_dc^{(d)}_{r_1}}\right)\right] + \sum_{s=1}^{S}m_{ts}\log \frac{c^{(dc)}_{r_0s}/c^{(d)}_{r_0}}{c^{(dc)}_{r_1s}/c^{(d)}_{r_1}}.   
\end{align}
For sufficiently concentrated rows $\bm{\kappa}_{r}$---i.e., $C_d$ is a good clustering using the majority vote estimator $\bm{\zeta}(C_d)$---we can let $c^{(dc)}_{rr}/c^{(d)}_{r}=1-\epsilon$ for small $\epsilon$ and $c^{(dc)}_{rs}/c^{(d)}_{r}=\epsilon/(S-1)$ for $r\neq s$. In this case, the expression simplifies to
\begin{align}
\Delta_t \approx \sum_{s=1}^{S}m_{ts}\log \frac{c^{(dc)}_{r_0s}/c^{(d)}_{r_0}}{c^{(dc)}_{r_1s}/c^{(d)}_{r_1}}
\approx (m_{tr_0}-m_{tr_1})\log \frac{(S-1)(1-\epsilon)}{\epsilon} > 0, 
\end{align} 
since $m_{tr_0}>m_{tr_1}$ by the nature of the majority vote estimator. Thus, the description length will increase from any local perturbation to the driver time series $\bm{\zeta}(C_d)$ when it is constructed as the majority vote estimator, so this driver is locally optimal for sufficiently densely sampled time series (i.e. $c_r^{(d)}\gg 1$) and sufficiently homogeneous clusterings (i.e. $c^{(dc)}_{rr}/c^{(d)}_{r}=1-\epsilon$ for small $\epsilon$).

\section{Details of case study results}
\label{app:details_main_text}
In this appendix we provide additional details for the case studies presented in the main text, including full cluster-level visualizations and summary statistics.

\begin{figure}[!htbp]
    \centering
    \includegraphics[width=0.9\columnwidth]{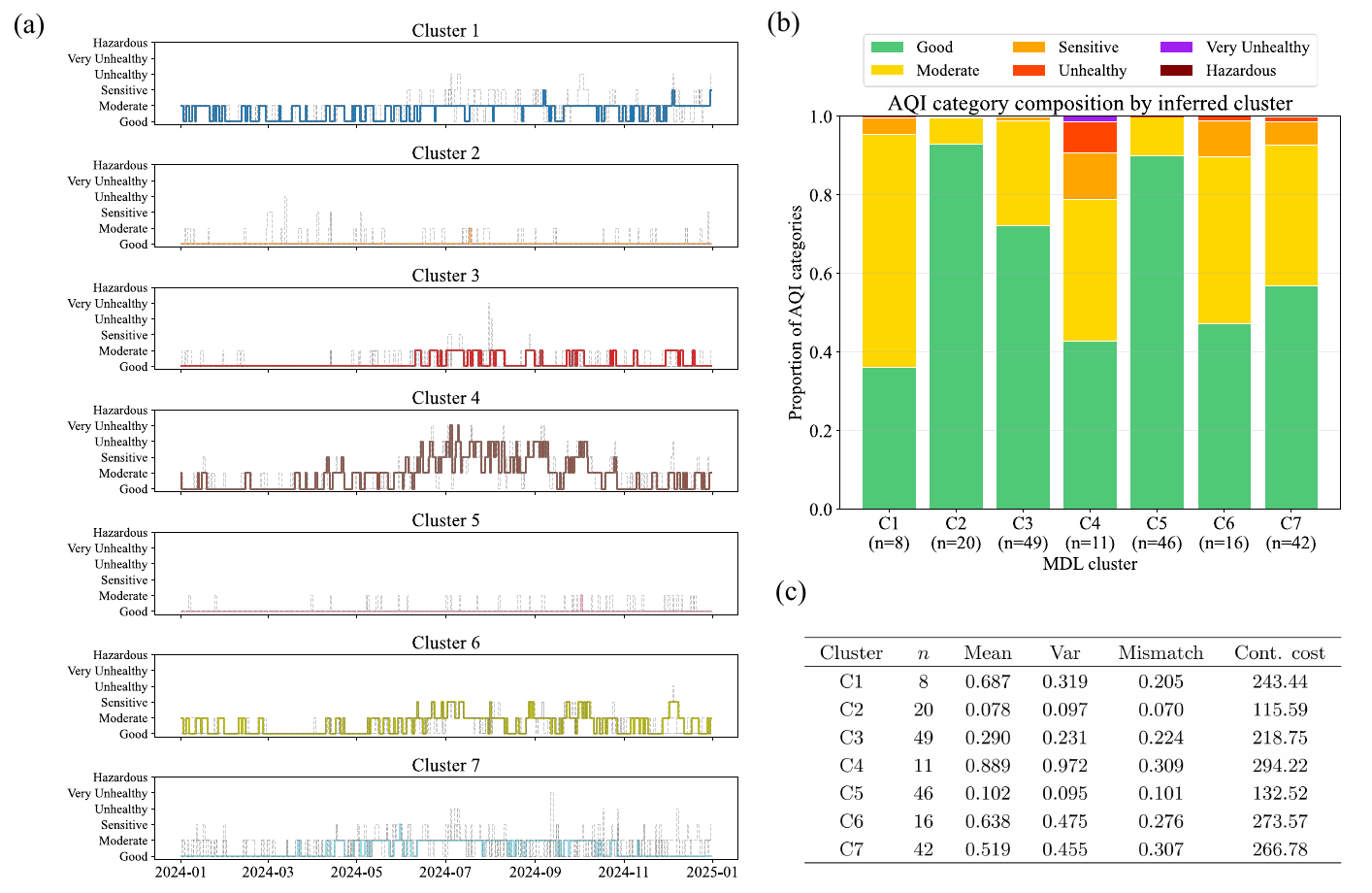}
    \caption{
    \textbf{Detailed cluster characterization for the California AQI dataset.}
    (a) Driver time series (colored) for all inferred clusters, with three representative member series (grey) per cluster.
    (b) Distribution of AQI categories within each cluster, aggregated over all sites and dates. Clusters are labeled as C1--C7.
    (c) Cluster-level summary statistics, including the number of sites $n$, mean and variance of AQI categories, mismatch rate (average fraction of time points where member series differ from the driver), and average contingency encoding cost (Eq.~(14) $\mathcal{L}(\bm{Z}\vert \mathcal{C})$) per series.
    }
    \label{fig:AQI_appendix}
\end{figure}

Fig.~\ref{fig:AQI_appendix} reports detailed results for the California AQI dataset. Panel (a) shows the driver time series for all inferred clusters together with representative member series. Panel (b) summarizes the distribution of AQI categories within each cluster. While these distributions capture overall air quality levels, they are not sufficient to uniquely determine cluster membership, as multiple clusters exhibit similar category proportions. Panel (c) reports cluster-level statistics, including size, mean, variance, mismatch rate, and average contingency encoding cost. Clusters with larger variability typically exhibit higher mismatch rates, leading to increased description length under the MDL objective.

\begin{figure}[!htbp]
    \centering
    \includegraphics[width=1\columnwidth]{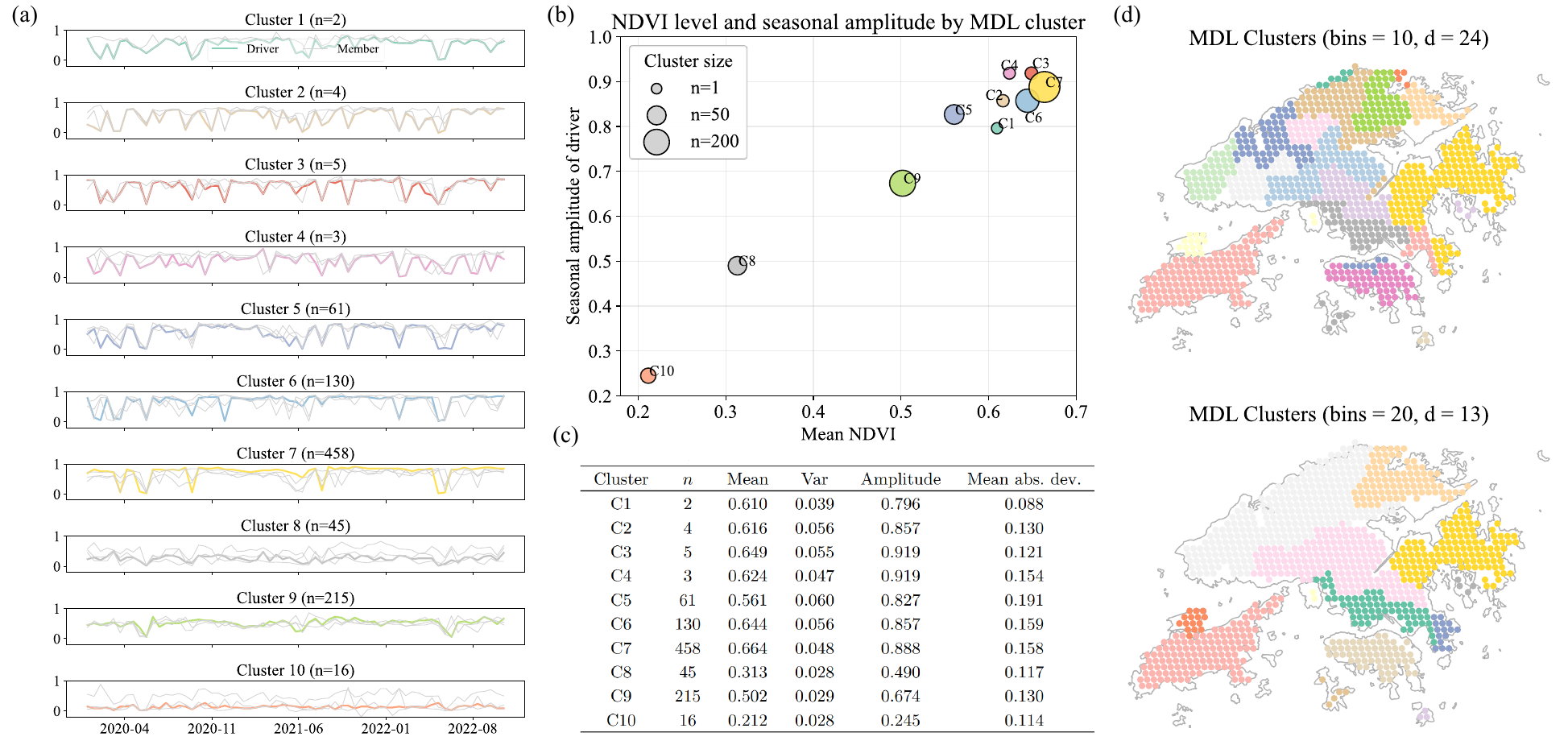}
    \caption{
    \textbf{Detailed cluster characterization for the Hong Kong NDVI dataset.}
    (a) Driver time series for all inferred clusters, with representative member series.
    (b) Each cluster is summarized by its mean NDVI level and seasonal amplitude (defined as the range of the driver time series), with marker size indicating the number of locations in the cluster.
    (c) Summary statistics for each cluster, including size, mean, variance, seasonal amplitude, and mean absolute deviation between member series and the cluster driver. 
    (d) Clustering results under different discretization levels ($S=10$ and $S=20$).
    }
    \label{fig:NDVI_appendix}
\end{figure}

Fig.~\ref{fig:NDVI_appendix} presents analogous results for the Hong Kong NDVI dataset. Panel (a) shows the driver and representative member time series for all clusters. Panel (b) visualizes cluster-level characteristics using mean NDVI and seasonal amplitude, highlighting clear differences between densely built, vegetated, and transitional regions. Panel (c) reports summary statistics including cluster size, mean, variance, seasonal amplitude, and average deviation from the driver. In panel (d) we show the effect of the discretization level. The main text reports the results at $S=40$, where the number of clusters stabilizes. At smaller values of $S$ (e.g., $S=10,20$), coarser discretization reduces similarity between time series, leading to more small clusters.

\section{Longitudinal result for AQI in California}
\label{app:longitudinal_AQI}
\begin{figure}[!htbp]
    \centering
    \includegraphics[width=0.85\columnwidth]{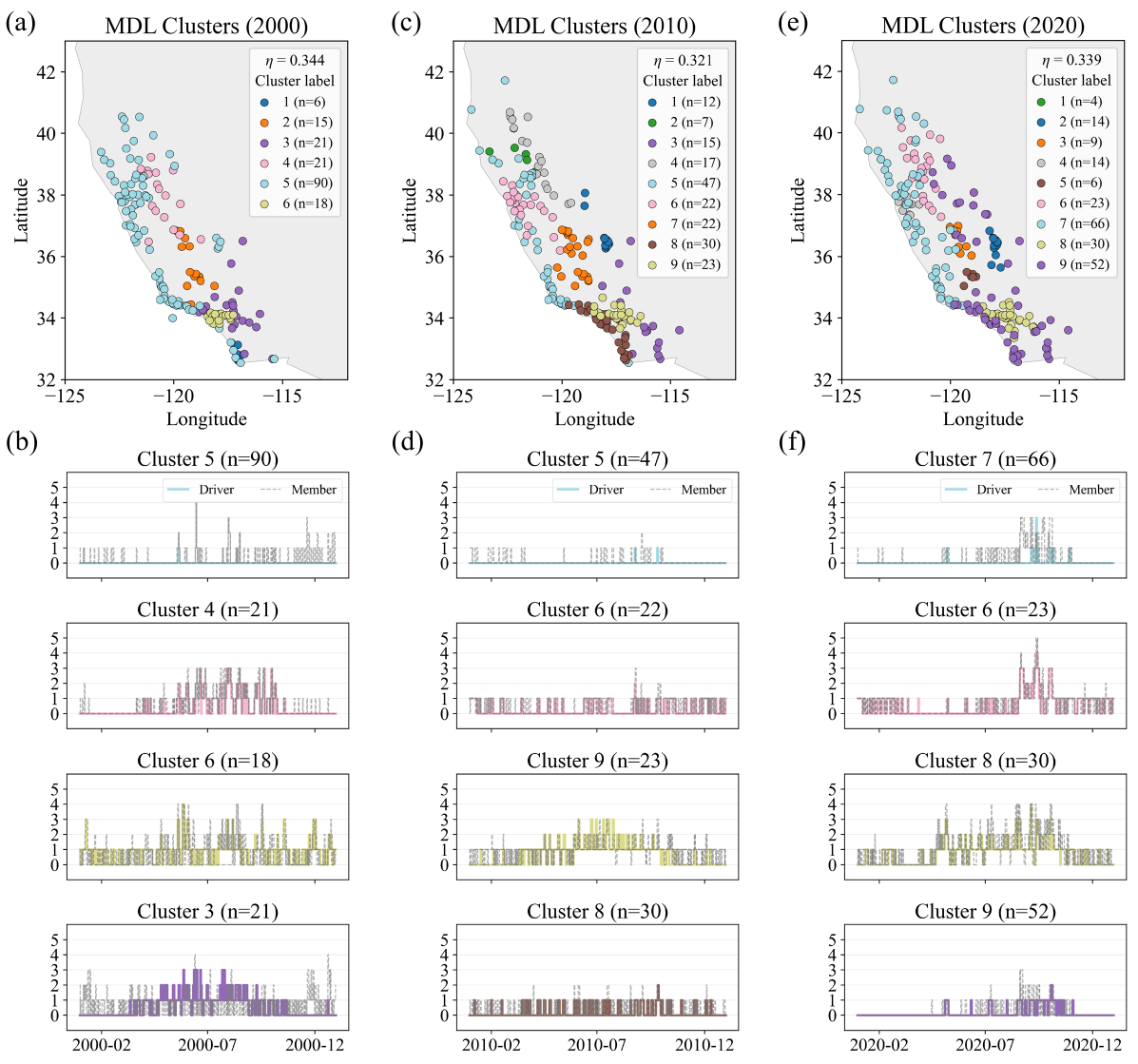}
    \caption{\textbf{Longitudinal regionalization results for California AQI.} (a)(c)(e) show the MDL-based spatial clustering results for 2000, 2010, and 2020, respectively. The legend in each map reports the inverse compression ratio $\eta$ and the cluster labels with the corresponding numbers of monitoring sites. (b)(d)(f) show the representative temporal patterns for the four largest clusters in each year, where the colored line denotes the inferred driver pattern and the gray lines denote member series. }
    \label{fig:AQI_different_years}
\end{figure}

Fig.~\ref{fig:AQI_different_years} presents a longitudinal comparison of the MDL-based regionalization results for California AQI in 2000, 2010, and 2020. While the clusters in different years are obtained from independent optimizations and are therefore not directly comparable, the figure reveals how the dominant spatial organization and temporal AQI patterns vary over time.
In 2000, the higher-pollution patterns appear more concentrated, with several of the largest clusters showing broader and more persistent pollution elevations. In 2010, the major patterns become more dispersed, and the temporal profiles of the largest clusters are generally weaker and flatter. In 2020, one large low-AQI cluster remains evident, while several other major clusters exhibit more distinct seasonal peaks, especially during summer.

\section{Larger-scale NDVI results}
\label{app:large_NDVI}
We further apply the MDL-based regionalization framework to larger-scale NDVI datasets in Guangdong Province and across mainland China. These examples complement the Hong Kong case study in the main text by showing that the method scales well to substantially larger spatial regions.

\begin{figure}[!htbp]
    \centering
    \includegraphics[width=1\columnwidth]{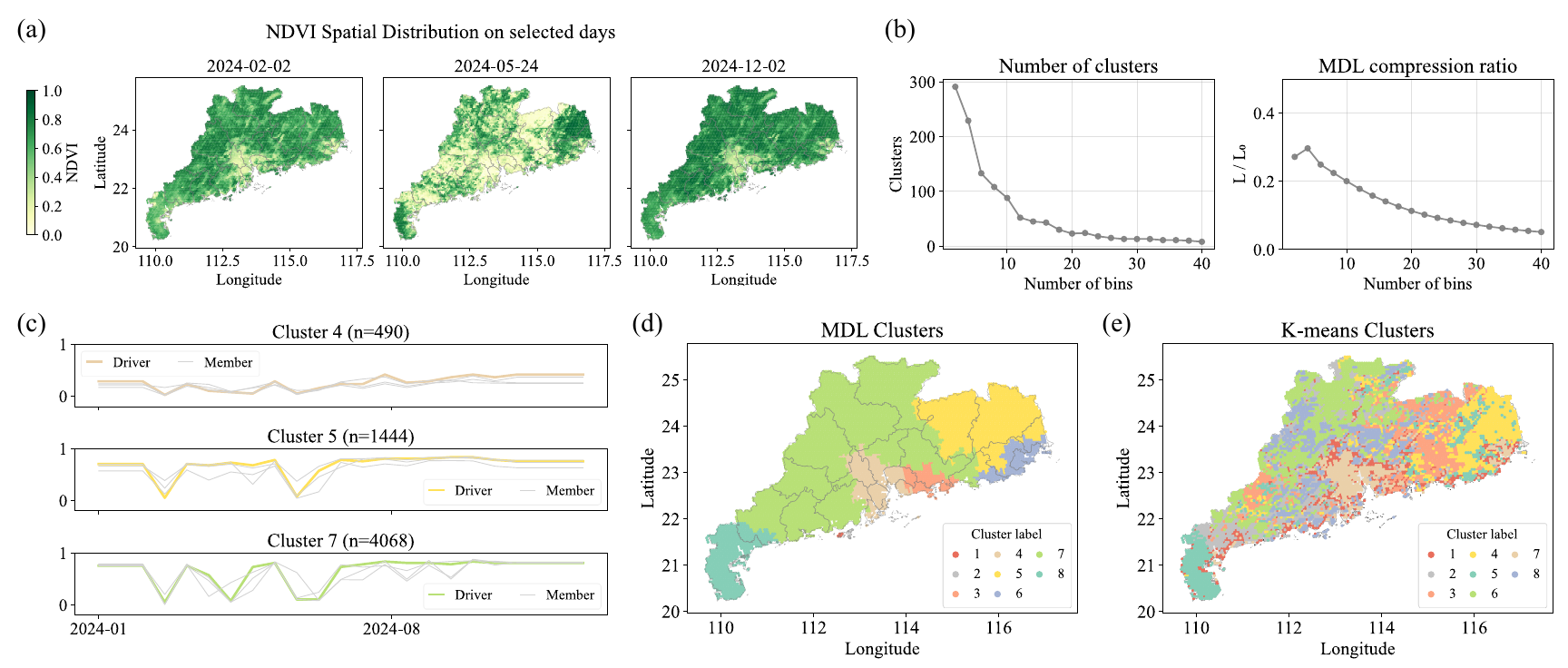}
    \caption{\textbf{Regionalization results of NDVI time series in Guangdong Province in 2024.} 
    The layout is the same as in the main text Fig.~4. The MDL result corresponds to $S=40$, for which the inferred number of clusters is 8. The gray lines in (d) show the administrative boundaries for reference.
    }
    \label{fig:NDVI_Guangdong}
\end{figure}

Fig.~\ref{fig:NDVI_Guangdong} shows the result for Guangdong in 2024. One large cluster covers much of western and central Guangdong, while the eastern part of the province forms another major region. Cluster 4 is concentrated around the Pearl River Delta and appears to capture its more urbanized core area. In the temporal patterns, Clusters 5 and 7 both exhibit persistently high NDVI with similar seasonal evolution, whereas Cluster 4 remains systematically lower across the year.

\begin{figure}[!htbp]
    \centering
    \includegraphics[width=0.85\columnwidth]{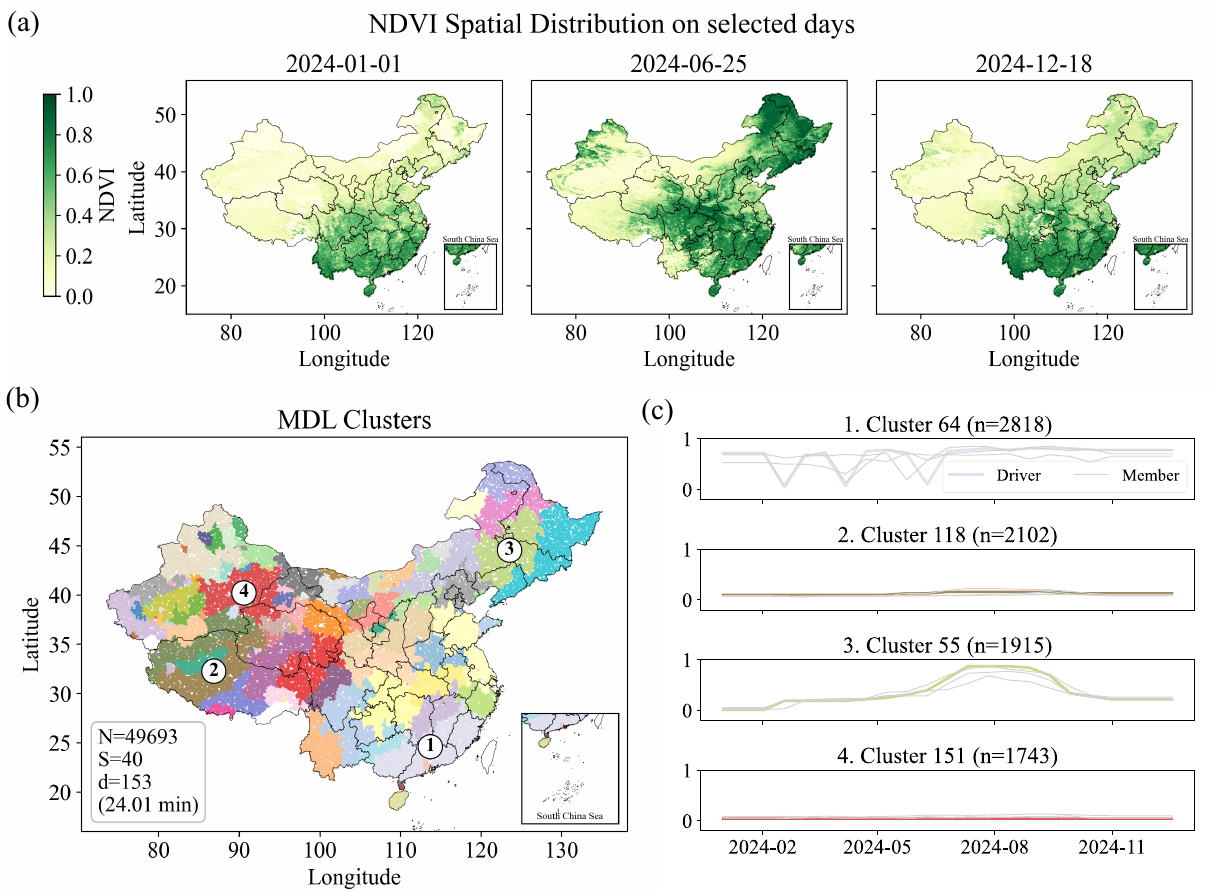}
    \caption{\textbf{Regionalization results of NDVI time series across China in 2024.}
    (a) Spatial distribution of NDVI on three selected dates. (b) Spatial clustering obtained by the MDL-based method at $S=40$, with 153 inferred clusters; the labels 1--4 mark the four largest regions. (c) Representative driver patterns and member series for these four largest regions. The calculation uses a sample of about 50 thousand sites and takes 21.19 minutes to run on a standard laptop using a pure Python implementation with no significant code optimization.}
    \label{fig:NDVI_China}
\end{figure}

At the national scale, Fig.~\ref{fig:NDVI_China} shows that the same framework can still extract interpretable regional structure from a much larger sample of about 50 thousand sites. The selected solution at $S=40$ contains 153 clusters. The four largest regions highlighted in Fig.~\ref{fig:NDVI_China}(b)--(c) illustrate several major vegetation regimes in China. Region 1 mainly covers southeastern China, including parts of Guangdong, Guangxi, Fujian, and Jiangxi, and shows relatively strong seasonal fluctuations. Region 3 extends across parts of Inner Mongolia and the northeastern provinces, with much higher NDVI during summer and autumn. Region 2 is concentrated over the Tibetan Plateau and Qinghai, while Region 4 is centered mainly in Xinjiang; both remain comparatively low throughout the year, consistent with sparse vegetation in these dry or high-altitude areas.

Overall, these larger-scale examples show that the MDL-based approach can recover spatially contiguous NDVI regions not only in compact urban territories, but also in large provincial and national domains. The resulting regions remain interpretable in terms of broad geographic and climatic contrasts, while the representative driver series provide a compact summary of their temporal vegetation dynamics.

\section{Algorithm time complexity}
\label{app:time_complexity}

In Supplementary Fig.~\ref{fig:runtime} we plot the log total runtime (in seconds) of the MDL regionalization algorithm against $\log N + \log \log N$, where $N$ is the number of spatial locations. The analysis is based on the NDVI dataset across China for the year 2024, using subsets of increasing size and running the full agglomerative clustering procedure.
The fitted slope is close to 1, indicating a scaling consistent with the theoretical $O(N \log N)$ complexity discussed in Sec.~\ref{sec:optimization}. This demonstrates that the method remains computationally efficient for large time series datasets.

\begin{figure}[!htbp]
    \centering
    \includegraphics[width=0.7\columnwidth]{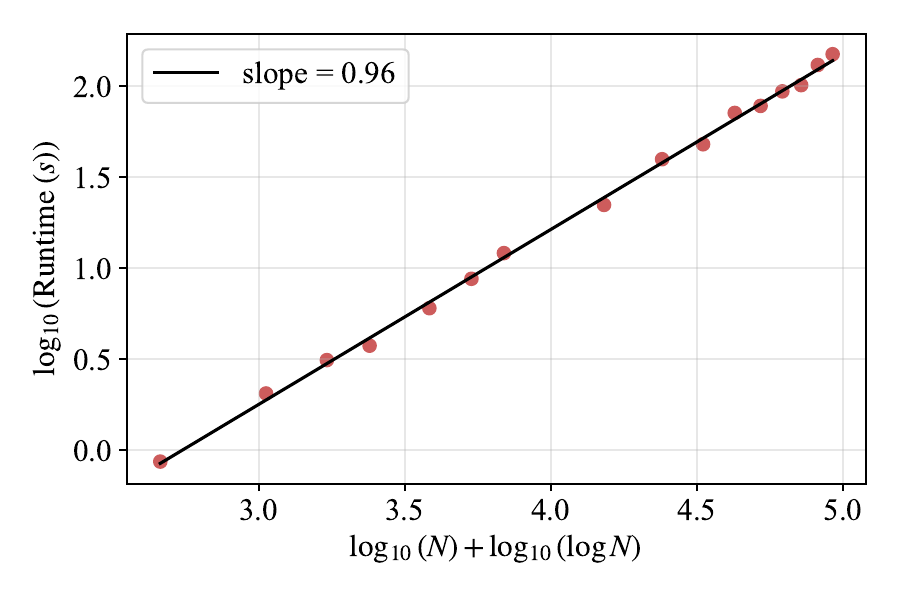}
    \caption{
    \textbf{Runtime scaling of the MDL regionalization algorithm.}
    Log total runtime (in seconds) versus $\log N + \log(\log N)$ for subsets of the China NDVI dataset (2024). The fitted slope ($\approx 0.96$) is close to 1, indicating near-linear scaling consistent with $O(N \log N)$ complexity.}
    \label{fig:runtime}
\end{figure}

\section{Comparison with exact solution}

\begin{figure}[!htbp]
    \centering
    \includegraphics[width=0.7\columnwidth]{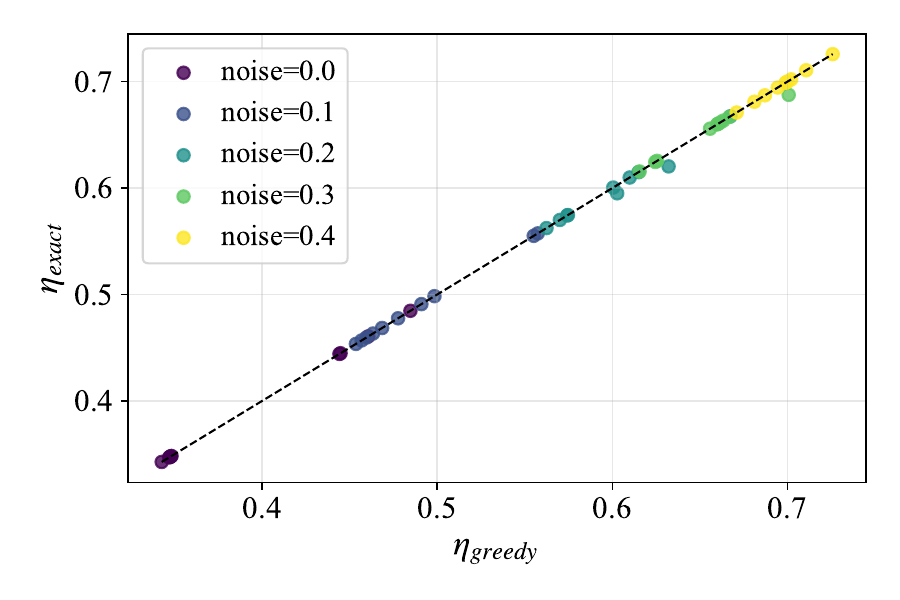}
    \caption{\textbf{Comparison between greedy and exact optimization of the description length objective.}
    The figure compares the inverse compression ratio $\eta$ obtained by the greedy algorithm and by exact enumeration on synthetic datasets with $N=10$ nodes. For each noise level, 10 simulations are performed, and different noise levels are indicated by different colors in the legend. The dashed line marks equality between the two methods.}
    \label{fig:compare_exact}
\end{figure}

To assess the quality of the greedy agglomerative optimization, we compare it with the exact solution on small synthetic datasets with $N=10$ spatial locations. The data are generated as in the synthetic experiments in the main text, except that some member series are additionally allowed to follow the driver pattern of another cluster, making the planted structure less clean and the problem more challenging. For each realization, we compute the optimal partition by exact enumeration over all contiguous partitions and by the greedy agglomerative algorithm.

We vary noise levels from 0 to 0.4, spanning settings in which the planted structure is progressively harder to recover. For each noise level, we generate 10 independent realizations and record the final inverse compression ratio $\eta$ for both methods.
As shown in Fig.~\ref{fig:compare_exact}, the greedy and exact results are usually identical. Only a few cases show small deviations, where the exact method yields a slightly smaller $\eta$. This suggests that the greedy algorithm works well for optimization in practice, at least for verifiable test cases in which exact enumeration is tractable. Meanwhile, it scales very well to large datasets, with a runtime scaling of only $O(N\log N)$ in the number of nodes $N$.

\end{document}